\documentclass{article}
\usepackage{arxiv}

\usepackage[utf8]{inputenc} 
\usepackage[T1]{fontenc}    
\usepackage{natbib}         
\usepackage{hyperref}       
\usepackage{url}            
\usepackage{booktabs}       
\usepackage{amsfonts}       
\usepackage{amsmath}        
\usepackage{amssymb}        
\usepackage{microtype}      
\usepackage{graphicx}
\usepackage{multirow}
\usepackage{adjustbox}   
\graphicspath{ {./} }

\usepackage{enumitem}

\newcommand{\Method}{HyperGuide}
\newcommand{\std}[1]{{\scriptsize$\pm$#1}} 

\title{HyperGuide: Hyperbolic Guidance for Efficient Multi-Step Reasoning in Large Language Models}

\author{
  Yuyu Liu \\
  Department of Computer Science \\
  Stony Brook University
  \And
  Haotian Xu \\
  Department of Applied Mathematics and Statistics \\
  Stony Brook University
  \And
  Yanan He \\
  Department of Computer Science \\
  Yale University
  \And
  Sarang Rajendra Patil \\
  Department of Data Science \\
  New Jersey Institute of Technology
  \And
  Mengjia Xu \\
  Department of Data Science \\
  New Jersey Institute of Technology
  \And
  Tengfei Ma \\
  Department of Biomedical Informatics \\
  Stony Brook University
}

\begin{document}

\maketitle

\begin{abstract}
Searching over alternative continuations lets large language models (LLMs) compare the downstream consequences of a reasoning step before committing to it, but expanding and evaluating many branches makes inference expensive. Training value models to score intermediate steps does not avoid this cost, as they still rank candidates at inference time. We introduce \Method{}, a framework that treats reasoning as movement through a tree of states embedded in hyperbolic space. \Method{} first trains a state encoder on the Poincar\'e ball, then trains a guidance head to predict the direction of a minimum-cost transition from the current state. The predicted direction is inserted into the decoder as a virtual token after each reasoning step, so inference follows a single autoregressive trajectory without expanding or reranking candidates. Because the supervision is ordinal rather than binary, it favors continuations that are both reliable and short, which steers generation toward successful, concise solutions. Experiments on competition mathematics and code generation show that \Method{} matches or exceeds the accuracy of search- and verifier-based baselines while generating substantially fewer tokens than these baselines and about as many as few-shot prompting.
\end{abstract}


\section{Introduction}\label{sec:introduction}

\begin{figure}[t]
    \centering
    \includegraphics[width=\linewidth]{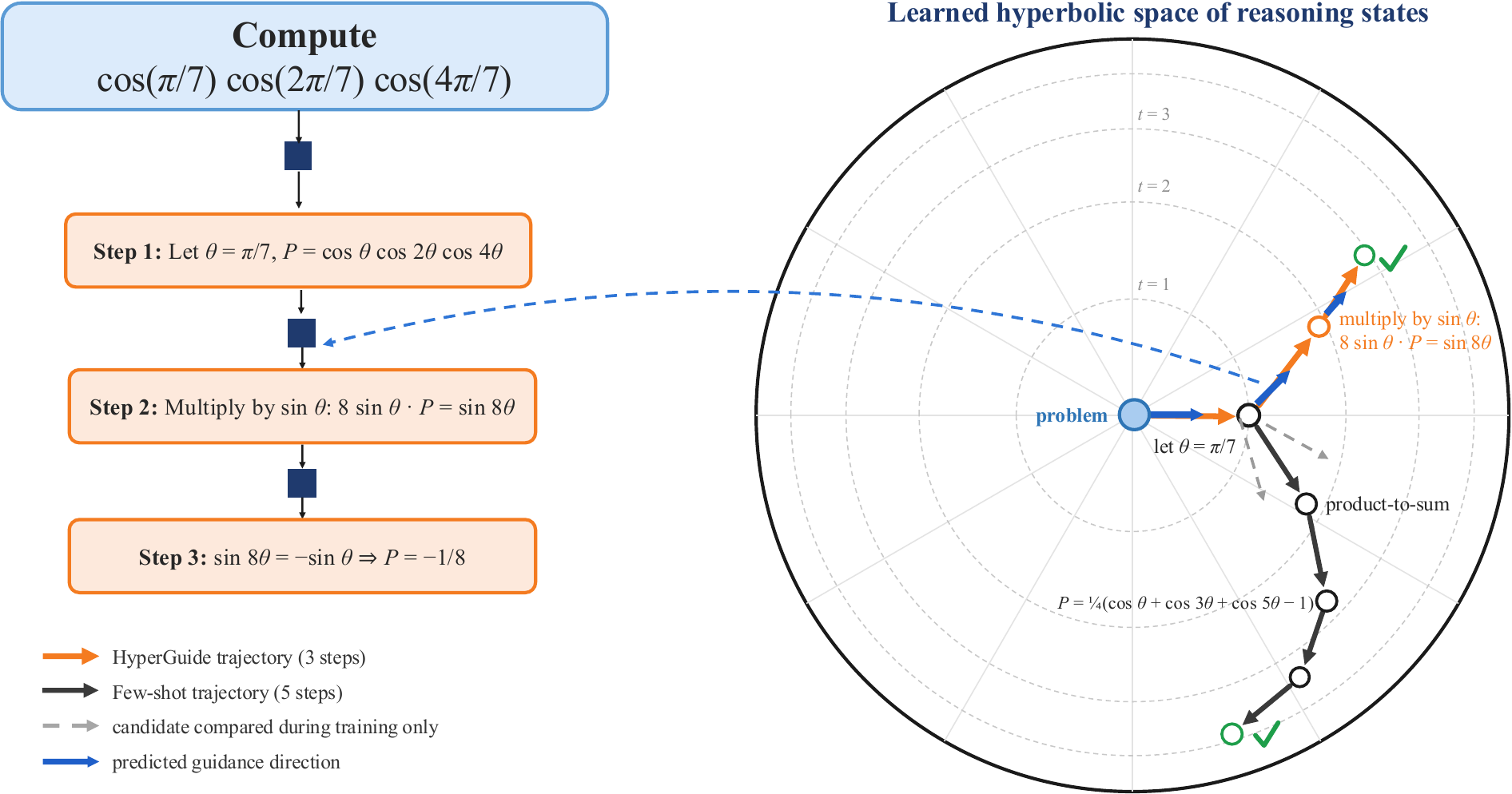}
    \caption{Reasoning example. \textbf{Left:} at inference time, each direction enters the decoder as a virtual token $\mathbf v_t$, and \Method{} generates a single trajectory. \textbf{Right:} reasoning states in the learned hyperbolic space, where distance from the problem root follows reasoning depth and different branches separate in direction. Dashed candidates are compared only during training, and \Method{} (orange) follows the predicted direction (blue arrows) of a lower-cost branch.}
    \label{fig:example}
\end{figure}

Large language models (LLMs) increasingly solve complex tasks through extended reasoning trajectories~\citep{chen_towards_2025,li_system_2025,sui_stop_2025}. Reasoning models such as DeepSeek-R1~\citep{deepseek-ai_deepseek-r1_2025} and s1~\citep{muennighoff_s1_2025} improve these trajectories through reasoning-focused training and additional deliberation at inference time. Generation nevertheless remains sequential: every next step commits to one continuation without comparing its downstream consequences against alternatives, and an unproductive choice may become apparent only several steps later~\citep{yang_beyond_2025,srivatsa_llms_2025,zhu_where_2025}. As shown in Figure~\ref{fig:example}, after setting $\theta=\pi/7$ to compute $\cos\theta\cos2\theta\cos4\theta$, applying product-to-sum identities is the habitual next step, yet it only turns the product into a sum of cosines that is no easier to evaluate, and the trajectory needs five steps to reach the answer.

Search-based methods such as Tree of Thoughts~\citep{yao_tree_2023} and Reasoning via Planning~\citep{hao_reasoning_2023} make this comparison explicit by expanding several continuations from an intermediate state and evaluating their outcomes. A second line of work learns value models and step-level verifiers that guide later decoding~\citep{liu_dont_2024,lightman_lets_2023,wang_math-shepherd_2024,yu_ovm_2024}. A learned scorer still needs several sampled candidates to rank at every step. Another line keeps a single decoding pass and intervenes in the model's continuous representations, feeding hidden states or soft thought tokens back into the decoder~\citep{hao_training_2025,shen_codi_2025,xu_softcot_2025,zhang_soft_2025}. The injected signal is not supervised by what happens after a step is taken.

At any point in reasoning, the same partial solution may admit many plausible continuations. These alternatives are not equally useful: some bring the reasoning closer to a successful solution, while others lead toward unproductive paths. This suggests viewing reasoning not merely as a linear sequence, but as movement through a structured space of possible continuations. If this structure can be captured geometrically, it may provide a local signal for choosing the next step without explicitly exploring the full tree.

Hyperbolic space is a natural fit for this branching structure. As reasoning proceeds, each partial solution may give rise to multiple continuations, causing the number of possible paths to grow rapidly with depth. Hyperbolic volume likewise grows exponentially with radius, providing increasing room to distinguish these alternatives as they branch~\citep{nickel_poincare_2017,nickel_learning_2018,sa_representation_2018}. At the same time, its metric can express the history shared by different continuations: those that diverge only recently can remain close, whereas those whose paths separated earlier can lie farther apart.

Building on this intuition, we propose \Method{} (\textbf{Hyper}bolic \textbf{Guid}anc\textbf{e} vector for LLM reasoning), which guides a single reasoning trajectory using directions learned in hyperbolic space. A state encoder maps representations produced by the frozen backbone into the Poincar'e ball. At each state encountered during training, we sample candidate next steps and rank their successor states by success-weighted continuation cost. A guidance head then learns, from the current state alone, to predict a direction aligned with one of the minimum-cost transitions, and the decoder learns to condition on this direction through a virtual token inserted after each completed step (Figure~\ref{fig:example}, left). We evaluate \Method{} on MATH-500 and a temporally held-out split of LiveCodeBench using two backbone families. Across these settings, it matches or exceeds the accuracy of search- and verifier-based baselines while generating roughly as many tokens as few-shot prompting. With Qwen2.5-7B-Instruct on MATH-500, for example, \Method{} achieves 77.5\% accuracy at 359 generated tokens, compared with 74.6\% at 4{,}754 tokens for value-model-guided search and 70.0\% at 527 tokens for few-shot prompting. Our contributions are summarized as follows:

\begin{enumerate}
    \item We formulate autoregressive reasoning as a prefix tree over partial trajectories and learn a hyperbolic representation of its states, where radius from the problem root tracks reasoning depth and pairwise distance reflects relative separation in the tree.
    \item We derive ordinal supervision over candidate next steps from final-answer verification alone, combining empirical continuation success and length, and use it to learn a guidance direction that enters the decoder as a virtual token, so inference follows a single trajectory without candidate expansion or reranking.
    \item Across math and code reasoning benchmarks with two backbone families, \Method{} matches or exceeds search- and verifier-based baselines at roughly one tenth of their generated tokens, with ablations confirming the roles of hyperbolic geometry and directional guidance.
\end{enumerate}

\section{Related Work}
\vspace{-0.4em}
\subsection{LLM Reasoning}
\vspace{-0.4em}
Single-pass prompting methods such as Chain-of-Thought~\citep{wei_chain--thought_2023}, Self-Consistency~\citep{wang_self-consistency_2023}, and Least-to-Most~\citep{zhou_least--most_2023} generate reasoning trajectories without explicitly comparing alternative future continuations at each step. Search-based methods such as Tree of Thoughts~\citep{yao_tree_2023} and Graph of Thoughts~\citep{besta_graph_2024} provide this lookahead by exploring multiple branches. A parallel line scores intermediate steps with learned verifiers~\citep{lightman_lets_2023, wang_math-shepherd_2024, uesato_solving_2022} to guide beam or best-of-$n$ decoding. Continuous-reasoning methods instead intervene in the model's representation space: Coconut~\citep{hao_training_2025} feeds the last hidden state back as the next input embedding, CODI~\citep{shen_codi_2025} compresses explicit chain-of-thought into continuous thoughts, and SoftCoT~\citep{xu_softcot_2025} injects soft thought tokens from a fixed assistant model.
\vspace{-0.4em}
\subsection{Learning from Search}
\vspace{-0.4em}
Search can provide supervision beyond final answers by exposing which intermediate decisions lead to successful continuations. Existing LLM methods use search trajectories for supervised fine-tuning or train separate value models to guide later search~\citep{liu_dont_2024,tian_toward_2024,zhang_rest-mcts_2024}. \Method{} instead converts verified continuations into ordinal supervision over candidate next steps and uses it to train guidance that is carried along a single autoregressive trajectory.
\vspace{-0.4em}
\subsection{Hyperbolic Representations}
\vspace{-0.4em}
Hyperbolic space embeds hierarchical or tree-like data with low distortion because ball volume grows exponentially with radius, matching tree branching. Poincar\'e~\citep{nickel_poincare_2017}, Lorentz~\citep{nickel_learning_2018}, and entailment-cone~\citep{ganea_hyperbolic_2018} embeddings demonstrate this empirically, with theoretical guarantees given by~\citet{sa_representation_2018}. A line of work generalizes standard neural-network layers to the Poincar\'e ball and other Riemannian manifolds~\citep{octavian-eugen_ganea_hyperbolic_2018, shimizu_hyperbolic_2021, chami_hyperbolic_2019, chen_fully_2022}, with applications in vision~\citep{khrulkov_hyperbolic_2020}, language~\citep{chen_probing_2021}, and recent transformer and LLM injections~\citep{yang_hyperbolic_2026}. \citet{raj_hyperbolic_2026} uses hyperbolic classifiers to probe reasoning-relevant hierarchy in frozen LLM hidden states. \Method{} instead learns the geometry of reasoning states and uses it to guide next-step generation.

\section{Preliminaries}\label{sec:preliminaries}
\vspace{-0.4em}
\paragraph{Reasoning states and continuation cost.}
A \emph{state} $s_t$ is the problem together with the reasoning steps generated so far. A next reasoning step $a_t$ produces the successor state $s_t\oplus a_t$, where $\oplus$ appends the step to the current state. We assume a verifier $\mathcal V$ that recognizes a correct final answer and a task-specific rule for identifying a completed reasoning step. The states reachable through continuations generated by a frozen backbone form a reasoning tree $\mathcal T$ rooted at the initial problem state $s_0$. Its depth is $t(s)=d_{\mathcal T}(s_0,s)$, and $d_{\mathcal T}(s_i,s_j)$ is the shortest-path distance between two states, both measured in reasoning steps. The subscript of $s_t$ denotes depth along a trajectory, whereas $s_i,s_j,s_k$ label arbitrary states of the same problem. The remaining reasoning cost $d(s)$ is the expected cost of a continuation sampled from the frozen backbone: a verified continuation incurs its number of logical reasoning steps, while an unsuccessful continuation incurs a fixed penalty $C_{\mathrm{fail}}$. We estimate this success-weighted cost from sampled continuations as $\hat d(s)$.
\vspace{-0.4em}
\paragraph{Hyperbolic geometry.}
\vspace{-0.4em}
We use the $n$-dimensional Poincar\'e ball $\mathbb D_c^n=\{\mathbf x\in\mathbb R^n:c\|\mathbf x\|^2<1\}$ with curvature $-c$, where $c>0$. Its origin is $\mathbf 0$, its geodesic distance is $d_{\mathbb D}$, and $T_{\mathbf z}\mathbb D_c^n$ denotes the tangent space at $\mathbf z$. The logarithmic map $\log^c_{\mathbf z}(\mathbf z')\in T_{\mathbf z}\mathbb D_c^n$ gives the tangent vector from $\mathbf z$ to $\mathbf z'$. The operator $\mathrm{PT}^c_{\mathbf x\to\mathbf y}:T_{\mathbf x}\mathbb D_c^n\to T_{\mathbf y}\mathbb D_c^n$ parallel-transports a tangent vector along the geodesic from $\mathbf x$ to $\mathbf y$. Both superscripts $c$ refer to the same curvature parameter. For tangent vectors at the same point, $\operatorname{sim}(\mathbf a,\mathbf b)$ denotes their cosine similarity.

\vspace{-0.4em}

\section{Methodology}\label{sec:methodology}
\begin{figure}[t]
    \centering
    \includegraphics[width=\linewidth]{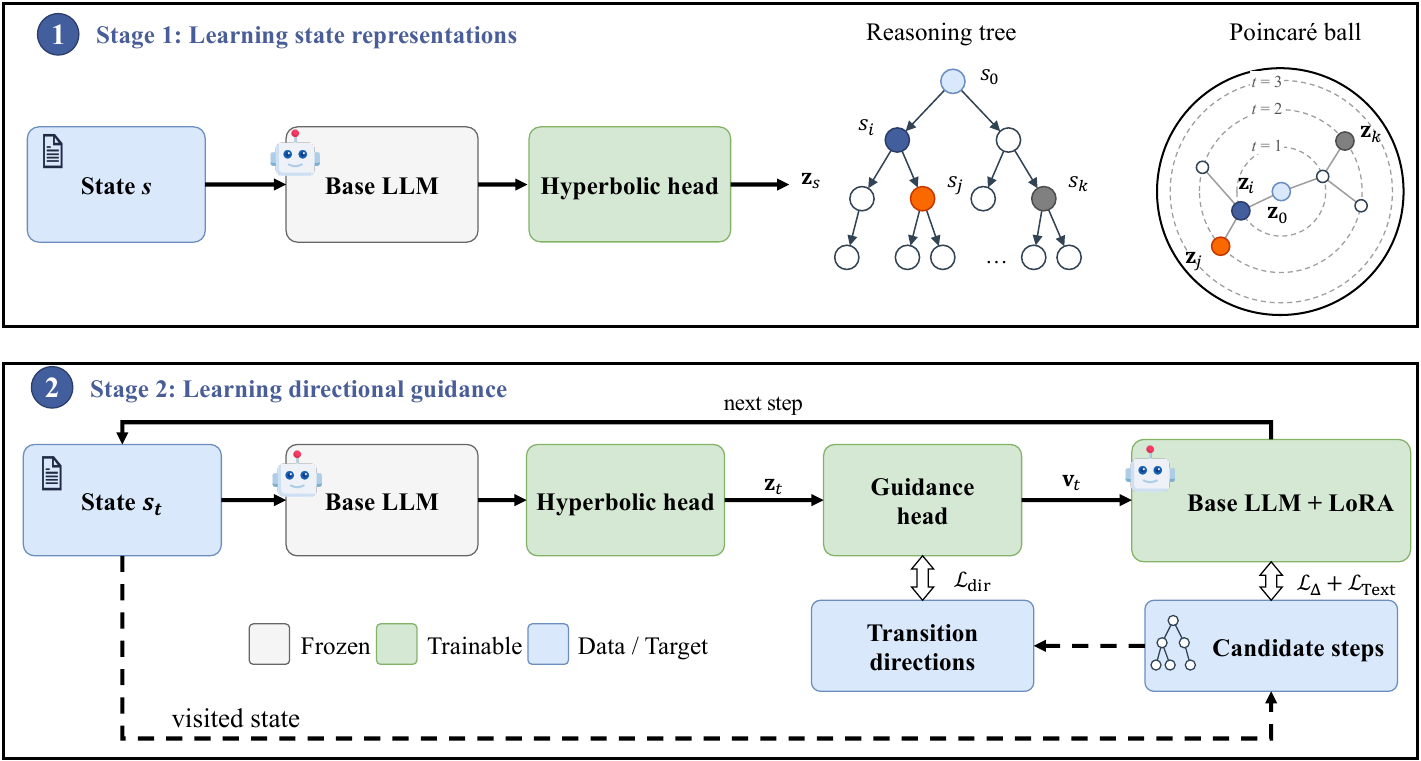}
    \caption{Overview of \Method{}. \textbf{Stage~1 (top):} a hyperbolic head maps frozen LLM representations of reasoning states into the Poincar\'e ball. \textbf{Stage~2 (bottom):} candidate steps ranked by continuation cost provide transition targets for the guidance head and preference supervision for the LoRA-adapted decoder.}
    \label{fig:pipeline}
\end{figure}

\subsection{Overview}

\Method{} is trained in two stages, as summarized in Figure~\ref{fig:pipeline}. Stage~1 learns hyperbolic representations of reasoning states (Section~\ref{ssec:training}). Stage~2 trains the model to predict a guidance direction from the current state and use it to condition next-step generation (Section~\ref{ssec:directional-guidance}). Section~\ref{ssec:inference} then describes how these components guide a single reasoning trajectory at inference time.

\subsection{Learning State Representations}\label{ssec:training}

In the first stage, we train an encoder $h_\phi$ to learn a representation for each reasoning state. Let $f_\theta(s)$ be the hidden representation of state $s$ from a frozen LLM backbone, and let $h_\phi$ map that representation into the Poincar\'e ball, giving the state embedding $\mathbf z_s = h_\phi(f_\theta(s)) \in \mathbb D_c^n$. Embeddings inherit the indices of their states, e.g., $\mathbf z_t=\mathbf z_{s_t}$ and $\mathbf z_i=\mathbf z_{s_i}$; in particular, $\mathbf z_0$ is the embedding of the problem root. Write $r(s)=d_{\mathbb D}(\mathbf z_0,\mathbf z_s)$ for the radius of $s$ from its root. For two states from the same problem with $t(s_i)<t(s_j)$ and a margin $\gamma_{\mathrm{rank}}>0$, the radial ranking objective is
\begin{equation}
    \mathcal L_{\mathrm{rank}}
    = \mathbb E\!\left[
    \max\!\left(0,
    r(s_i)-r(s_j)
    + \gamma_{\mathrm{rank}}\bigl(t(s_j)-t(s_i)\bigr)
    \right)\right].
    \label{eq:stage1-rank-main}
\end{equation}
Thus deeper states are trained to lie farther from the root of their reasoning tree.

We also preserve relative distances in the reasoning tree. For triplets $(s_i,s_j,s_k)$ from the same problem satisfying $d_{\mathcal T}(s_i,s_j)<d_{\mathcal T}(s_i,s_k)$ and a margin $\gamma_{\mathrm{metric}}>0$, let
\begin{equation}
    \mathcal L_{\mathrm{metric}}
    = \mathbb E\!\left[
    w_{ijk}\max\!\left(0,
    d_{\mathbb D}(\mathbf z_i,\mathbf z_j)
    - d_{\mathbb D}(\mathbf z_i,\mathbf z_k)
    + \gamma_{\mathrm{metric}}
    \right)\right],
    \label{eq:stage1-metric-main}
\end{equation}
where the importance weight $w_{ijk}$ down-weights triplets with large tree distances. This keeps nearby reasoning states closer in the learned geometry than states on more distant branches. The complete Stage~1 objective is
\begin{equation}
    \mathcal L_{\mathrm{Geom}}
    = \mathcal L_{\mathrm{rank}}
    + \lambda_{\mathrm{metric}}\mathcal L_{\mathrm{metric}},
    \label{eq:stage1_loss}
\end{equation}
where $\lambda_{\mathrm{metric}}\geq 0$ weights the tree-metric objective.

\subsection{Learning Directional Guidance}\label{ssec:directional-guidance}

In the second stage, starting from these state representations, we train a guidance head to predict a direction for the next reasoning step. The predicted direction is converted into a virtual token, and we train the decoder to favor lower-cost next steps when the token is present.

\paragraph{Guidance token construction.} Consider a state $s_t$ visited during training. We form a candidate set $\mathcal C(s_t)=\{a_t^{(i)}\}_i$, where each next reasoning step gives a successor state $s_{t+1}^{(i)}=s_t\oplus a_t^{(i)}$. The continuation estimator assigns each successor a remaining-cost estimate $\hat d_i=\hat d(s_{t+1}^{(i)})$. We embed each successor and express its transition from the current state in the tangent space at $\mathbf z_t$ as $\mathbf u_t^{(i)}=\log^c_{\mathbf z_t}(\mathbf z_{t+1}^{(i)})\in T_{\mathbf z_t}\mathbb D_c^n$, the geometric direction taken by candidate step $i$. A guidance head predicts a direction $\hat{\mathbf u}_t = q_\omega(\mathbf z_t,\mathbf z_0)$ from the current state and its problem root. Because $\hat{\mathbf u}_t$ lies in a state-specific tangent space, we parallel-transport it to the tangent space at the origin and map it to the backbone hidden width:
\begin{equation}
    \mathbf v_t = g_\psi\!\left(\mathrm{PT}^c_{\mathbf z_t\to\mathbf 0}(\hat{\mathbf u}_t)\right)
    = g_\psi\!\left(\mathrm{PT}^c_{\mathbf z_t\to\mathbf 0}\bigl(q_\omega(\mathbf z_t,\mathbf z_0)\bigr)\right).
    \label{eq:virtual-token}
\end{equation}
The resulting $\mathbf v_t$ is the virtual guidance token inserted into the decoder.

\paragraph{Guidance objective.} We order candidate transitions by successor cost, $\mathcal R(s_t)=\{(i,j):\hat d_i<\hat d_j\}$, excluding ties, and collect the minimum-cost candidates $\mathcal P(s_t)=\{i:\hat d_i=\min_k \hat d_k\}$. We train the predicted direction to align with at least one member of $\mathcal P(s_t)$, with a margin $\gamma_{\mathrm{dir}}>0$:
\begin{equation}
    \mathcal L_{\mathrm{dir}}
    = \mathbb E_{s_t,j\notin\mathcal P(s_t)}\!\left[
    \max\!\left(0,
    \operatorname{sim}(\hat{\mathbf u}_t,\mathbf u_t^{(j)})
    -\max_{i\in\mathcal P(s_t)}\operatorname{sim}(\hat{\mathbf u}_t,\mathbf u_t^{(i)})
    +\gamma_{\mathrm{dir}}
    \right)\right].
    \label{eq:direction-set-loss}
\end{equation}
For $(i,j)\in\mathcal R(s_t)$, define
\begin{equation}
\begin{aligned}
    \Delta_g^{ij}
    &= \ell(a_t^{(i)}\mid s_t,\mathbf v_t)-\ell(a_t^{(j)}\mid s_t,\mathbf v_t),\\
    \Delta_0^{ij}
    &= \ell(a_t^{(i)}\mid s_t,\varnothing)-\ell(a_t^{(j)}\mid s_t,\varnothing),
\end{aligned}
\label{eq:margins}
\end{equation}
where $\ell$ is sequence log-likelihood under the same adapted decoder, and $\varnothing$ denotes that no virtual guidance token is inserted. With logistic sigmoid $\sigma(x)=1/(1+e^{-x})$, we optimize
\begin{equation}
    \mathcal L_{\mathrm{Align}}
    = \mathcal L_{\mathrm{dir}}+\lambda_{\Delta}\mathcal L_{\Delta},
    \qquad
    \mathcal L_{\Delta}
    = -\mathbb E_{s_t,(i,j)\in\mathcal R(s_t)}
    \left[\log\sigma\!\left(\Delta_g^{ij}-\Delta_0^{ij}\right)\right].
\label{eq:guide-loss}
\end{equation}
Here $\lambda_{\Delta}\geq 0$ weights the guidance-preference term. The directional loss selects one minimum-cost transition, while $\mathcal L_{\Delta}$ requires the virtual token to increase the decoder's relative preference for lower-cost transitions. The latter ordering reflects both empirical continuation success and the lengths of successful continuations.

We additionally use an auxiliary text loss $\mathcal L_{\mathrm{Text}}$ on minimum-cost candidate steps and verified final-answer steps while retaining the state-geometry objective (Appendix~\ref{app:text-loss}). The complete Stage~2 objective is
\begin{equation}
    \mathcal L_{\mathrm{Guide}}
    = \mathcal L_{\mathrm{Align}}+\lambda_{\mathrm{text}}\mathcal L_{\mathrm{Text}},
    \label{eq:stage2-loss}
\end{equation}
where $\lambda_{\mathrm{text}}\geq 0$ weights the auxiliary training term.

\subsection{Inference}\label{ssec:inference}

At inference time, \Method{} generates a single autoregressive trajectory without using continuation costs. After each completed reasoning step, the accumulated state $s_t$ is encoded as $\mathbf z_t$, and the guidance head predicts a direction from $\mathbf z_t$ and the problem-root embedding $\mathbf z_0$. This direction is converted into the virtual token $\mathbf v_t$ by Equation~\ref{eq:virtual-token} and appended to the decoder stream before generation continues. The process repeats until the model produces a final answer. No candidate branches are sampled, scored, or reranked at inference time.

\section{Experiments}\label{sec:experiments}

\subsection{Experimental Settings}

\paragraph{Implementation Details.} We instantiate \Method{} on \textbf{Qwen2.5-7B-Instruct}~\citep{qwen_qwen25_2025} and \textbf{Llama-3.1-8B-Instruct}~\citep{aimeta_llama_2024}, two model families at a similar scale. Stage~1 trains the state encoder on top of the frozen backbone. Stage~2 jointly refines the state encoder and trains the guidance head, the up-projector, and a LoRA adapter on the decoder. We train \Method{} and every ablation variant with three random seeds (4, 44, and 444). Hyperparameters and other implementation details are provided in Appendix~\ref{sec:additional-exp-settings}.

\paragraph{Baselines.} We compare \Method{} with two types of baselines. \textit{Training-free} methods include \textbf{Few-shot} prompting~\citep{brown_language_2020}, \textbf{Self-Consistency}~\citep{wang_self-consistency_2023} (majority voting over sampled trajectories, MATH-500 only), \textbf{Tree of Thoughts}~\citep{yao_tree_2023} (search over candidate steps scored by the backbone), and \textbf{Soft Thinking}~\citep{zhang_soft_2025} (continuous concept tokens in a single pass). \textit{Training-based} methods include \textbf{PT-SFT}~\citep{wang_guiding_2024} (fine-tuning with learned planning tokens), \textbf{OVM}~\citep{yu_ovm_2024} (beam search guided by an outcome value model), and the latent-reasoning methods \textbf{Coconut}~\citep{hao_training_2025} and \textbf{SoftCoT}~\citep{xu_softcot_2025}.

\paragraph{Datasets and Metrics.} We evaluate on all 500 problems in \textbf{MATH-500}~\citep{lightman_lets_2023} and on 341 code-generation problems in a temporally held-out \textbf{LiveCodeBench} test window~\citep{jain_livecodebench_2025}. The LiveCodeBench split is defined by release date so that every test problem postdates the training and validation problems (Table~\ref{tab:dataset_splits}). We report two metrics over each complete test set: (1) \textit{Accuracy} on MATH-500 and \textit{Pass@1} on LiveCodeBench, the percentage of problems with a verified final answer or passing code; and (2) \textit{Tokens}, the mean number of generated tokens. For seed-dependent methods, we report the mean and standard deviation over the three seeds.

\subsection{Main Results}\label{ssec:main-results}

\begin{table}[t]
\centering
\footnotesize
\setlength{\tabcolsep}{2pt}
\caption{Main results on MATH-500 and the temporally held-out LiveCodeBench code-generation split, with Qwen2.5-7B-Instruct and Llama-3.1-8B-Instruct as backbones. Tokens is the mean number of generated tokens. Self-Consistency is evaluated on MATH-500 only. The best result in each column is bolded and second best is underlined.}
\label{tab:main}
\begin{adjustbox}{max width=\linewidth}
\begin{tabular}{lcccccccc}
\toprule
 & \multicolumn{4}{c}{MATH-500} & \multicolumn{4}{c}{LiveCodeBench} \\
\cmidrule(lr){2-5}\cmidrule(lr){6-9}
 & \multicolumn{2}{c}{Qwen2.5-7B} & \multicolumn{2}{c}{Llama-3.1-8B} & \multicolumn{2}{c}{Qwen2.5-7B} & \multicolumn{2}{c}{Llama-3.1-8B} \\
\cmidrule(lr){2-3}\cmidrule(lr){4-5}\cmidrule(lr){6-7}\cmidrule(lr){8-9}
Method & Acc. $\uparrow$ & Tokens $\downarrow$ & Acc. $\uparrow$ & Tokens $\downarrow$ & Pass@1 $\uparrow$ & Tokens $\downarrow$ & Pass@1 $\uparrow$ & Tokens $\downarrow$ \\
\midrule
\multicolumn{9}{l}{\textit{Training-free}} \\
\quad Few-shot         &70.0 &527.4 &43.4 &325.8 &22.9 &376.8 &11.1 &319.9 \\
\quad Self-Consistency &71.3\std{0.6} &2504.3\std{57.3} &\underline{46.7}\std{0.7} &1549.0\std{16.1} & -- & -- & -- & -- \\
\quad Tree of Thoughts &66.2\std{2.5} &5532.0\std{94.6} &40.2\std{1.3} &3859.4\std{96.4} &20.9\std{1.8} &3965.2\std{85.2} &11.3\std{1.1} &3441.5\std{16.7} \\
\quad Soft Thinking    &71.0 &506.4 &44.0 &313.6 &21.1 &\underline{357.1} &15.2 &307.2 \\
\midrule
\multicolumn{9}{l}{\textit{Training-based}} \\
\quad PT-SFT           &69.5\std{1.4} &532.1\std{6.5} &41.1\std{1.6} &337.9\std{2.0} &19.3\std{1.9} &379.1\std{8.8} &13.0\std{2.1} &313.0\std{4.3} \\
\quad OVM              &\underline{74.6}\std{1.7} &4754.3\std{27.4} &46.4\std{1.1} &3968.2\std{22.6} &\textbf{24.9}\std{1.5} &3716.7\std{87.2} &\underline{17.8}\std{1.8} &3118.5\std{19.7} \\
\quad Coconut          &26.2\std{0.7} &\textbf{36.0}\std{0.3} &17.5\std{0.5} &\textbf{25.3}\std{0.6} &3.0\std{0.4} &\textbf{196.3}\std{4.4} &2.1\std{0.3} &\textbf{159.2}\std{3.9} \\
\quad SoftCoT          &70.8\std{0.8} &528.9\std{7.7} &45.3\std{1.8} &322.0\std{6.3} &23.2\std{1.5} &371.6\std{7.6} &13.6\std{1.1} &319.7\std{4.4} \\
\addlinespace[2pt]
\quad \textbf{\Method{} (ours)} &\textbf{77.5}\std{1.4} &\underline{358.9}\std{7.1} &\textbf{48.6}\std{0.9} &\underline{310.8}\std{4.5} &\underline{24.4}\std{0.7} &365.7\std{7.8} &\textbf{19.3}\std{0.9} &\underline{306.7}\std{7.2} \\
\bottomrule
\end{tabular}
\end{adjustbox}
\end{table}

Table~\ref{tab:main} reports the main results. \Method{} attains the highest mean accuracy in three of the four settings. On MATH-500, it leads the strongest baseline by more than one standard deviation of either method; on LiveCodeBench, the gap between \Method{} and OVM is within one standard deviation of OVM with both backbones. 

\vspace{-0.4em}
\paragraph{Comparison with search- and verifier-based methods.} OVM is the strongest baseline, but it scores several candidate steps at every state and generates 10 to 13 times as many tokens as \Method{}. \Method{} exceeds its accuracy by 2.2 to 2.9 points on MATH and is on par with it on LiveCodeBench, while following a single trajectory. Tree of Thoughts generates 11 to 15 times as many tokens and still falls below Few-shot in three of the four settings and only matches it in the fourth, which suggests that the self-evaluation of a small backbone is not reliable enough.

\vspace{-0.4em}
\paragraph{Comparison with single-pass methods.} Soft Thinking, SoftCoT, and PT-SFT also decode one trajectory, but they stay within 4.1 points of Few-shot. Coconut replaces most textual steps with latent ones and produces very short outputs, and its accuracy falls far below Few-shot on both benchmarks. \Method{} exceeds the best single-pass baseline by 3.3 to 6.5 points in three settings; on LiveCodeBench with Qwen2.5-7B, its 1.2-point lead over SoftCoT is within the standard deviation of SoftCoT. The difference lies in what supervises the injected signal: the planning tokens and soft thoughts of these baselines are not trained on what happens after a step is taken, whereas the guidance token is trained to raise the decoder's preference for the successor with the lower remaining cost.

\subsection{Ablations}\label{ssec:ablation}

We ablate the components behind the two parts of \Method{}. Each variant changes one component and keeps the rest of the training and inference procedure fixed. Table~\ref{tab:ablation} reports the results. Several additional ablations are reported in Appendix~\ref{app:additional-ablation}.

\begin{table}[t]
\centering
\footnotesize
\caption{Ablations on Qwen2.5-7B-Instruct. The best result in each column is bolded.}
\label{tab:ablation}
\begin{tabular}{lcccc}
\toprule
& \multicolumn{2}{c}{MATH-500} & \multicolumn{2}{c}{LiveCodeBench} \\
\cmidrule(lr){2-3}\cmidrule(lr){4-5}
Variant & Accuracy $\uparrow$ & Tokens $\downarrow$ & Pass@1 $\uparrow$ & Tokens $\downarrow$ \\
\midrule
\textbf{\Method{}} &\textbf{77.5}\std{1.4} &\textbf{358.9}\std{7.1} &\textbf{24.4}\std{0.7} &\textbf{365.7}\std{7.8} \\
\quad Euclidean geometry &74.4\std{1.5} &366.7\std{6.3} &20.0\std{1.8} &378.2\std{7.6} \\
\quad w/o $\mathcal L_{\mathrm{rank}}$ &72.6\std{1.2} &453.7\std{4.7} &21.6\std{1.9} &392.5\std{6.4} \\
\quad w/o $\mathcal L_{\mathrm{metric}}$ &74.2\std{1.6} &362.1\std{2.2} &22.2\std{1.4} &371.2\std{8.8} \\
\quad w/o guidance token &70.4\std{0.7} &377.4\std{2.6} &20.8\std{0.8} &370.3\std{3.3} \\
\bottomrule
\end{tabular}
\end{table}

\textit{Euclidean geometry.} In this variant, we replace the Poincar\'e ball with $\mathbb R^n$ and keep both Stage~1 objectives, with Euclidean distances in place of hyperbolic ones. Accuracy drops by 3.1 points on MATH and Pass@1 by 4.4 points on LiveCodeBench, the largest drop on that benchmark, while the token counts change by less than 4\%. The same supervision is therefore less effective in flat space. 

\textit{w/o $\mathcal L_{\mathrm{rank}}$.} This variant removes the radial depth-ranking objective, so distance from the problem root is no longer explicitly ordered by reasoning depth. Accuracy drops by 4.9 points on MATH-500 and Pass@1 by 2.8 points on LiveCodeBench. It also clearly lengthens trajectories, from 358.9 to 453.7 tokens on MATH-500 and from 365.7 to 392.5 on LiveCodeBench.

\textit{w/o $\mathcal L_{\mathrm{metric}}$.} This variant removes the tree-metric objective, so pairwise distances are no longer trained to follow branch structure. Accuracy drops by 3.3 points on MATH-500 and Pass@1 by 2.2 points on LiveCodeBench, and the token counts stay within 2\% of the full model. Branch structure thus mainly affects whether the model reaches a correct solution, not how long that solution is. 

\textit{w/o guidance token.} In this variant, we train the decoder adapter on the same data but neither predict nor insert $\mathbf v_t$, so the decoder receives no signal from the learned geometry. This yields the largest accuracy drop on MATH-500, 7.1 points, and a 3.6-point drop in Pass@1 on LiveCodeBench. The gain of \Method{} therefore does not come from adapter training alone: the learned representation improves generation only when its direction is passed to the decoder.

\subsection{Representation of Reasoning Depth and Prefix-Tree Structure}\label{ssec:mechanism-geometry}
\vspace{-0.4em}
The ablations show that the two Stage~1 objectives matter for accuracy, but not whether the learned space has the structure they are meant to impose. We therefore test the deployed state encoder on problems it has never seen: does radius from the problem root follow reasoning depth, and does pairwise distance follow distance in the reasoning tree?

\paragraph{Setup.} On Qwen2.5-7B-Instruct and MATH-500, we build a reasoning tree per evaluation problem with the training-time continuation sampling, which yields held-out states with depth $t(s)$, tree distances $d_{\mathcal T}$, and candidate steps with costs $\hat d$. We encode every state with the \Method{} encoder and with the \emph{Euclidean geometry} encoder of Table~\ref{tab:ablation}, both trained with seed 4. Two metrics mirror the two objectives: \emph{depth order}, the share of same-problem pairs with $t(s_i)<t(s_j)$ for which $r(s_i)<r(s_j)$, and \emph{triplet accuracy}, the share of triplets with $d_{\mathcal T}(s_i,s_j)<d_{\mathcal T}(s_i,s_k)$ whose embedding distances are ordered the same way; chance is 50\% for both. We report triplet accuracy over all triplets and over the subset whose states lie on different branches, where the metric tests branch separation rather than step counting along one path.

\begin{table}[t]
\centering
\footnotesize
\caption{Structure of the learned space on held-out MATH-500 states (Qwen2.5-7B-Instruct), in \%. Chance is 50\% for every column.}
\label{tab:geometry}
\begin{tabular}{lccc}
\toprule
& & \multicolumn{2}{c}{Triplet accuracy $\uparrow$} \\
\cmidrule(lr){3-4}
Encoder & Depth order $\uparrow$ & All & Across branches \\
\midrule
Hyperbolic (\Method{}) &94.5 &88.7 &82.1 \\
Euclidean &87.3 &78.2 &69.8 \\
\bottomrule
\end{tabular}
\end{table}

\begin{figure}[t]
    \centering
    \includegraphics[width=0.9\linewidth]{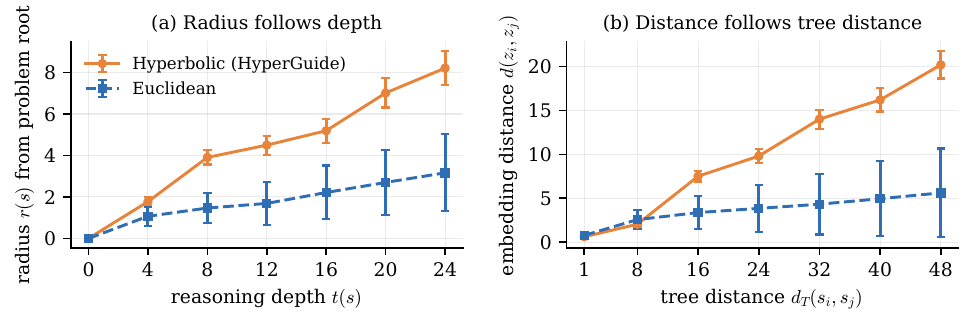}
    \vspace{-0.4em}
    \caption{Learned geometry on held-out MATH-500 states. \textbf{(a)} Mean radius from the problem root at each reasoning depth. \textbf{(b)} Mean embedding distance between two states at each tree distance. Error bars show one standard deviation. The hyperbolic encoder keeps both relations monotone; the Euclidean encoder compresses deep states and distant branches.}
    \label{fig:geometry}
\end{figure}

\paragraph{Radius follows depth.} As shown in Table~\ref{tab:geometry}, both encoders order most held-out pairs by depth (94.5\% against 87.3\%), since $\mathcal L_{\mathrm{rank}}$ trains this directly. The difference appears at larger depth (Figure~\ref{fig:geometry}(a)): in the hyperbolic encoder the mean radius keeps growing with $t(s)$, whereas in the Euclidean encoder it flattens, so deep states from the same problem become hard to tell apart by radius alone.

\paragraph{Distance follows tree distance.} Triplet accuracy separates the two geometries more clearly: 88.7\% against 78.2\% over all triplets, and the gap widens to 12.3 points (82.1\% against 69.8\%) on triplets across branches. In flat space, states that diverged several steps ago are placed at nearly the same distance as states that diverged recently (Figure~\ref{fig:geometry}(b)), while the hyperbolic encoder keeps the distance growing with $d_{\mathcal T}$.
\vspace{-0.4em}

\subsection{Causal Effect of the Guidance Token on Next-Step Selection}\label{ssec:mechanism-token}
\vspace{-0.4em}
Section~\ref{ssec:mechanism-geometry} does not show whether the guidance head can select a lower-cost branch from the current state, or whether the decoder uses the resulting token. We therefore keep the trained \Method{} model fixed and change only the token it receives at inference time.
\vspace{-0.4em}
\paragraph{Setup.} We compare four conditions on Qwen2.5-7B-Instruct and MATH-500: \emph{Predicted} inserts $\mathbf v_t$; \emph{No token} inserts nothing; \emph{Shuffled} inserts the token predicted at a random state of a different test problem, which is in-distribution but carries the direction of an unrelated state; \emph{Negated} replaces $\hat{\mathbf u}_t$ with $-\hat{\mathbf u}_t$ in Equation~\ref{eq:virtual-token}. We report accuracy and tokens as in Table~\ref{tab:main}, and two step-level metrics on the held-out states of Section~\ref{ssec:mechanism-geometry} with at least two candidate steps of different cost, where the decoder only scores fixed candidates. \emph{Pair shift} is the share of pairs $(i,j)\in\mathcal R(s_t)$ with $\Delta_g^{ij}>\Delta_0^{ij}$, that is, the token moves likelihood toward the lower-cost candidate; chance is 50\%. \emph{Min-cost top-1} is the share of states at which the most likely candidate has minimum cost.
\vspace{-0.4em}
\begin{table}[t]
\centering
\footnotesize
\caption{Inference-time interventions on the guidance token of the trained \Method{} model (Qwen2.5-7B-Instruct, MATH-500). Pair shift and Min-cost top-1 are in \% and are computed on held-out states with fixed candidate steps. Pair shift compares against the same decoder without a token, so it is undefined for \emph{No token}. The best result in each column is bolded.}
\label{tab:token-intervention}
\begin{tabular}{lcccc}
\toprule
& \multicolumn{2}{c}{Decoding} & \multicolumn{2}{c}{Step level} \\
\cmidrule(lr){2-3}\cmidrule(lr){4-5}
Inserted token & Accuracy $\uparrow$ & Tokens $\downarrow$ & Pair shift $\uparrow$ & Min-cost top-1 $\uparrow$ \\
\midrule
Predicted $\mathbf v_t$ (\Method{}) &\textbf{77.4} &\textbf{352.7} &\textbf{77.8} &\textbf{61.8} \\
No token &70.2 &380.0 &-- &45.2 \\
Shuffled &71.4 &389.2 &52.0 &47.6 \\
Negated &65.2 &405.3 &34.4 &40.8 \\
\bottomrule
\end{tabular}
\end{table}
\vspace{-0.4em}
\begin{figure}[t]
    \centering
    \includegraphics[width=\linewidth]{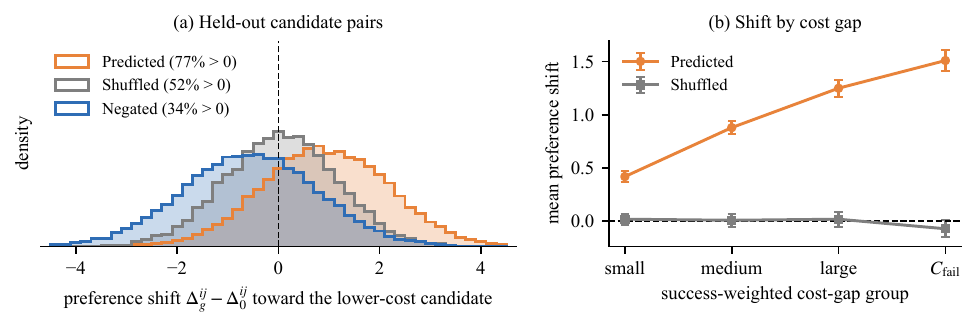}
    \vspace{-0.4em}
    \caption{Effect of the inserted token on the decoder's preference between two candidate steps at held-out MATH-500 states. The preference shift $\Delta_g^{ij}-\Delta_0^{ij}$ is positive when the token moves likelihood toward the lower-cost candidate. \textbf{(a)} Distribution over all candidate pairs for the three inserted tokens. \textbf{(b)} Mean shift with 95\% confidence intervals, grouped by success-weighted cost gap; $C_{\mathrm{fail}}$ denotes pairs in which the higher-cost candidate has no verified continuation.}
    \label{fig:token-effect}
\end{figure}

\paragraph{The decoder reads the content of the token.} With the predicted token, the likelihood margin moves toward the lower-cost candidate in 77.8\% of the pairs, and the most likely candidate is a minimum-cost one at 61.8\% of the states, against 45.2\% without a token. A shuffled token has no such effect. Its pair shift stays near chance (52.0\%), its distribution in Figure~\ref{fig:token-effect}(a) is centered near zero, and its accuracy differs by 1.2 points from decoding without a token. The shuffled token therefore does not reproduce the effect of the predicted token. The 7-point accuracy difference between the predicted token and \emph{No token} depends on carrying a direction specific to the current state.
\vspace{-0.4em}
\paragraph{The effect follows the cost gap.} Figure~\ref{fig:token-effect}(b) groups the pairs by the difference in remaining cost between the two candidates. The shift caused by the predicted token is smallest for small finite cost gaps, grows with the gap, and is largest when the higher-cost candidate has no verified continuation. The token therefore acts most strongly on the decisions that matter most, and it also separates candidates that both lead to a solution, which binary supervision cannot do.
\vspace{-0.4em}
\section{Conclusion}
\vspace{-0.4em}
This work introduced \Method{}, which represents the branching structure of reasoning in hyperbolic space and uses it to guide a single autoregressive trajectory. By learning from alternative continuations during training, \Method{} avoids expanding or reranking candidate steps at inference time. Across MATH-500 and LiveCodeBench with two backbone families, it matches or exceeds the accuracy of search- and verifier-based baselines while using roughly one tenth of their generated tokens. Ablations and inference-time interventions further show that both the learned geometry and the resulting directional guidance contribute to these gains. More broadly, our results suggest that the geometry of reasoning prefixes provides a useful coordinate system for guiding generation.


\subsubsection*{Reproducibility Statement}

Source code for training and evaluation is available at
\url{https://github.com/yuyuliu11037/HyperGuide}. Model architecture, training objectives, hyperparameters, and dataset construction are described in Section~\ref{sec:methodology} and Appendices~\ref{sec:additional-exp-settings} and~\ref{sec:training-details}. All seed-dependent results are reported as the mean and standard deviation over seeds 4, 44, and 444.


\bibliographystyle{plainnat}
\bibliography{references}

@inproceedings{hao_training_2025,
	title={Training Large Language Models to Reason in a Continuous Latent Space},
author={Shibo Hao and Sainbayar Sukhbaatar and DiJia Su and Xian Li and Zhiting Hu and Jason E Weston and Yuandong Tian},
booktitle={Second Conference on Language Modeling},
year={2025}
}

@inproceedings{shen_codi_2025,
	title = {{CODI}: Compressing chain-of-thought into continuous space via self-distillation},
	author = {Shen, Zhenyi and Yan, Hanqi and Zhang, Linhai and Hu, Zhanghao and Du, Yali and He, Yulan},
	booktitle = {Proceedings of the 2025 Conference on Empirical Methods in Natural Language Processing},
	pages = {677--693},
	year = {2025},
}

@inproceedings{xu_softcot_2025,
	title = {{SoftCoT}: Soft chain-of-thought for efficient reasoning with {LLMs}},
	author = {Xu, Yige and Guo, Xu and Zeng, Zhiwei and Miao, Chunyan},
	booktitle = {Proceedings of the 63rd Annual Meeting of the Association for Computational Linguistics (Volume 1: Long Papers)},
	pages = {23336--23351},
	year = {2025},
}

@inproceedings{raj_hyperbolic_2026,
	title = {Hyperbolic geometry of reasoning: Probing {LLM} hidden states},
	author = {Raj, Arnav},
	booktitle = {ICLR 2026 Workshop on Geometry-grounded Representation Learning and Generative Modeling (GRaM), Tiny Paper Track},
	year = {2026},
}

@inproceedings{yu_ovm_2024,
	title = {{OVM}, outcome-supervised value models for planning in mathematical reasoning},
	author = {Yu, Fei and Gao, Anningzhe and Wang, Benyou},
	booktitle = {Findings of the Association for Computational Linguistics: NAACL 2024},
	pages = {858--875},
	year = {2024},
}

@article{qwen_qwen25_2025,
	title = {{Qwen2.5} technical report},
	author = {Qwen and Yang, An and Yang, Baosong and Zhang, Beichen and Hui, Binyuan and Zheng, Bo and Yu, Bowen and Li, Chengyuan and Liu, Dayiheng and Huang, Fei and Wei, Haoran and Lin, Huan and Yang, Jian and Tu, Jianhong and Zhang, Jianwei and Yang, Jianxin and Yang, Jiaxi and Zhou, Jingren and Lin, Junyang and Dang, Kai and Lu, Keming and Bao, Keqin and Yang, Kexin and Yu, Le and Li, Mei and Xue, Mingfeng and Zhang, Pei and Zhu, Qin and Men, Rui and Lin, Runji and Li, Tianhao and Tang, Tianyi and Xia, Tingyu and Ren, Xingzhang and Ren, Xuancheng and Fan, Yang and Su, Yang and Zhang, Yichang and Wan, Yu and Liu, Yuqiong and Cui, Zeyu and Zhang, Zhenru and Qiu, Zihan},
	journal = {arXiv preprint arXiv:2412.15115},
	year = {2024},
}

@inproceedings{brown_language_2020,
	title={Language models are few-shot learners},
  author={Brown, Tom and Mann, Benjamin and Ryder, Nick and Subbiah, Melanie and Kaplan, Jared D and Dhariwal, Prafulla and Neelakantan, Arvind and Shyam, Pranav and Sastry, Girish and Askell, Amanda and others},
  booktitle={Advances in Neural Information Processing Systems},
  volume={33},
  pages={1877--1901},
  year={2020}
}

@inproceedings{yang_hyperbolic_2026,
	title     = {Hyperbolic Fine-Tuning for Large Language Models},
  author    = {Menglin Yang and Ram Samarth B B and Aosong Feng and Bo Xiong and Jiahong Liu and Irwin King and Rex Ying},
  booktitle = {Advances in Neural Information Processing Systems},
  volume    = {38},
  year      = {2025}
}

@inproceedings{shimizu_hyperbolic_2021,
	title={Hyperbolic Neural Networks++},
author={Ryohei Shimizu and YUSUKE Mukuta and Tatsuya Harada},
booktitle={International Conference on Learning Representations},
year={2021}
}

@inproceedings{zhang_rest-mcts_2024,
	title={Rest-mcts*: Llm self-training via process reward guided tree search},
  author={Zhang, Dan and Zhoubian, Sining and Hu, Ziniu and Yue, Yisong and Dong, Yuxiao and Tang, Jie},
  booktitle={Advances in Neural Information Processing Systems},
  volume={37},
  pages={64735--64772},
  year={2024}
}

@inproceedings{tian_toward_2024,
	title={Toward self-improvement of llms via imagination, searching, and criticizing},
  author={Tian, Ye and Peng, Baolin and Song, Linfeng and Jin, Lifeng and Yu, Dian and Han, Lei and Mi, Haitao and Yu, Dong},
  booktitle={Advances in Neural Information Processing Systems},
  volume={37},
  pages={52723--52748},
  year={2024}
}

@inproceedings{liu_dont_2024,
	title = {Don't throw away your value model! {G}enerating more preferable text with value-guided {Monte-Carlo} tree search decoding},
	author = {Liu, Jiacheng and Cohen, Andrew and Pasunuru, Ramakanth and Choi, Yejin and Hajishirzi, Hannaneh and Celikyilmaz, Asli},
	booktitle = {Conference on Language Modeling},
	year = {2024},
}

@article{uesato_solving_2022,
	title = {Solving math word problems with process- and outcome-based feedback},
	author = {Uesato, Jonathan and Kushman, Nate and Kumar, Ramana and Song, Francis and Siegel, Noah and Wang, Lisa and Creswell, Antonia and Irving, Geoffrey and Higgins, Irina},
	journal = {arXiv preprint arXiv:2211.14275},
	year = {2022},
}

@inproceedings{wang_math-shepherd_2024,
	title={Math-shepherd: Verify and reinforce llms step-by-step without human annotations},
  author={Wang, Peiyi and Li, Lei and Shao, Zhihong and Xu, Runxin and Dai, Damai and Li, Yifei and Chen, Deli and Wu, Yu and Sui, Zhifang},
  booktitle={Proceedings of the 62nd Annual Meeting of the Association for Computational Linguistics (Volume 1: Long Papers)},
  pages={9426--9439},
  year={2024}
}

@inproceedings{besta_graph_2024,
	title = {{Graph of Thoughts}: Solving elaborate problems with large language models},
	author = {Besta, Maciej and Blach, Nils and Kubicek, Ales and Gerstenberger, Robert and Podstawski, Michal and Gianinazzi, Lukas and Gajda, Joanna and Lehmann, Tomasz and Niewiadomski, Hubert and Nyczyk, Piotr and Hoefler, Torsten},
	booktitle = {Proceedings of the {AAAI} Conference on Artificial Intelligence},
	volume = {38},
	pages = {17682--17690},
	year = {2024},
}

@inproceedings{hao_reasoning_2023,
	title = {Reasoning with language model is planning with world model},
	author = {Hao, Shibo and Gu, Yi and Ma, Haodi and Hong, Joshua and Wang, Zhen and Wang, Daisy and Hu, Zhiting},
	booktitle = {Proceedings of the 2023 Conference on Empirical Methods in Natural Language Processing},
	pages = {8154--8173},
	year = {2023},
}

@article{deepseek-ai_deepseek-r1_2025,
	title = {{DeepSeek-R1} incentivizes reasoning in {LLMs} through reinforcement learning},
	author = {Guo, Daya and Yang, Dejian and Zhang, Haowei and Song, Junxiao and Wang, Peiyi and others},
	journal = {Nature},
	volume = {645},
	number = {8081},
	pages = {633--638},
	year = {2025},
}

@inproceedings{octavian-eugen_ganea_hyperbolic_2018,
	title = {Hyperbolic neural networks},
	author = {Ganea, Octavian-Eugen and Bécigneul, Gary and Hofmann, Thomas},
	booktitle = {Advances in Neural Information Processing Systems},
	volume = {31},
	year = {2018},
}

@inproceedings{wang_guiding_2024,
	title = {Guiding language model reasoning with planning tokens},
	author = {Wang, Xinyi and Caccia, Lucas and Ostapenko, Oleksiy and Yuan, Xingdi and Wang, William Yang and Sordoni, Alessandro},
	booktitle = {Conference on Language Modeling},
	year = {2024},
}

@inproceedings{khrulkov_hyperbolic_2020,
	title = {Hyperbolic image embeddings},
	author = {Khrulkov, Valentin and Mirvakhabova, Leyla and Ustinova, Evgeniya and Oseledets, Ivan and Lempitsky, Victor},
	booktitle = {Proceedings of the {IEEE/CVF} Conference on Computer Vision and Pattern Recognition},
	pages = {6417--6427},
	year = {2020},
}

@inproceedings{muennighoff_s1_2025,
	title={s1: Simple test-time scaling},
  author={Muennighoff, Niklas and Yang, Zitong and Shi, Weijia and Li, Xiang Lisa and Fei-Fei, Li and Hajishirzi, Hannaneh and Zettlemoyer, Luke and Liang, Percy and Cand{\`e}s, Emmanuel and Hashimoto, Tatsunori B},
  booktitle={Proceedings of the 2025 Conference on Empirical Methods in Natural Language Processing},
  pages={20286--20332},
  year={2025}
}

@inproceedings{lightman_lets_2023,
	title = {Let's verify step by step},
	author = {Lightman, Hunter and Kosaraju, Vineet and Burda, Yura and Edwards, Harri and Baker, Bowen and Lee, Teddy and Leike, Jan and Schulman, John and Sutskever, Ilya and Cobbe, Karl},
	booktitle = {International Conference on Learning Representations},
	year = {2024},
}

@inproceedings{zhou_least--most_2023,
	title = {Least-to-most prompting enables complex reasoning in large language models},
	author = {Zhou, Denny and Schärli, Nathanael and Hou, Le and Wei, Jason and Scales, Nathan and Wang, Xuezhi and Schuurmans, Dale and Cui, Claire and Bousquet, Olivier and Le, Quoc V. and Chi, Ed H.},
	booktitle = {International Conference on Learning Representations},
	year = {2023},
}

@inproceedings{wang_self-consistency_2023,
	title = {Self-consistency improves chain of thought reasoning in language models},
	author = {Wang, Xuezhi and Wei, Jason and Schuurmans, Dale and Le, Quoc V. and Chi, Ed H. and Narang, Sharan and Chowdhery, Aakanksha and Zhou, Denny},
	booktitle = {International Conference on Learning Representations},
	year = {2023},
}

@inproceedings{yao_tree_2023,
	title = {{Tree of Thoughts}: Deliberate problem solving with large language models},
	author = {Yao, Shunyu and Yu, Dian and Zhao, Jeffrey and Shafran, Izhak and Griffiths, Thomas L. and Cao, Yuan and Narasimhan, Karthik},
	booktitle = {Advances in Neural Information Processing Systems},
	volume = {36},
	pages = {11809--11822},
	year = {2023},
}

@inproceedings{wei_chain--thought_2023,
	title={Chain-of-thought prompting elicits reasoning in large language models},
  author={Wei, Jason and Wang, Xuezhi and Schuurmans, Dale and Bosma, Maarten and Xia, Fei and Chi, Ed and Le, Quoc V and Zhou, Denny and others},
  booktitle={Advances in Neural Information Processing Systems},
  volume={35},
  pages={24824--24837},
  year={2022}
}

@article{aimeta_llama_2024,
	title = {The {Llama} 3 herd of models},
	author = {Grattafiori, Aaron and Dubey, Abhimanyu and Jauhri, Abhinav and Pandey, Abhinav and Kadian, Abhishek and others},
	journal = {arXiv preprint arXiv:2407.21783},
	year = {2024},
}

@inproceedings{chen_probing_2021,
	title = {Probing {BERT} in hyperbolic spaces},
	author = {Chen, Boli and Fu, Yao and Xu, Guangwei and Xie, Pengjun and Tan, Chuanqi and Chen, Mosha and Jing, Liping},
	booktitle = {International Conference on Learning Representations},
	year = {2021},
}

@inproceedings{nickel_learning_2018,
	title={Learning continuous hierarchies in the {L}orentz model of hyperbolic geometry},
  author={Nickel, Maximilian and Kiela, Douwe},
  booktitle={International conference on machine learning},
  pages={3779--3788},
  year={2018},
  organization={PMLR}
}

@inproceedings{sa_representation_2018,
	title={Representation tradeoffs for hyperbolic embeddings},
  author={Sala, Frederic and De Sa, Chris and Gu, Albert and R{\'e}, Christopher},
  booktitle={International conference on machine learning},
  pages={4460--4469},
  year={2018},
  organization={PMLR}
}

@inproceedings{ganea_hyperbolic_2018,
	title={Hyperbolic entailment cones for learning hierarchical embeddings},
  author={Ganea, Octavian and B{\'e}cigneul, Gary and Hofmann, Thomas},
  booktitle={International conference on machine learning},
  pages={1646--1655},
  year={2018},
  organization={PMLR}
}

@inproceedings{nickel_poincare_2017,
	title={Poincar{\'e} embeddings for learning hierarchical representations},
  author={Nickel, Maximilian and Kiela, Douwe},
  booktitle={Advances in Neural Information Processing Systems},
  volume={30},
  year={2017}
}

@inproceedings{chen_fully_2022,
	title={Fully hyperbolic neural networks},
  author={Chen, Weize and Han, Xu and Lin, Yankai and Zhao, Hexu and Liu, Zhiyuan and Li, Peng and Sun, Maosong and Zhou, Jie},
  booktitle={Proceedings of the 60th Annual Meeting of the Association for Computational Linguistics (Volume 1: Long Papers)},
  pages={5672--5686},
  year={2022}
}

@inproceedings{chami_hyperbolic_2019,
	title={Hyperbolic graph convolutional neural networks},
  author={Chami, Ines and Ying, Zhitao and R{\'e}, Christopher and Leskovec, Jure},
  booktitle={Advances in Neural Information Processing Systems},
  volume={32},
  year={2019}
}

@inproceedings{yang_beyond_2025,
	title={Beyond the First Error: Process Reward Models for Reflective Mathematical Reasoning.},
  author={Yang, Zhaohui and He, Chenghua and Shi, Xiaowen and Deng, Shihong and Li, Linjing and Yin, Qiyue and Jiang, Daxin},
  booktitle={EMNLP (Findings)},
  pages={4711--4728},
  year={2025}
}

@inproceedings{srivatsa_llms_2025,
	title={LLMs cannot spot math errors, even when allowed to peek into the solution},
  author={Srivatsa, Kv Aditya and Maurya, Kaushal Kumar and Kochmar, Ekaterina},
  booktitle={Proceedings of the 2025 Conference on Empirical Methods in Natural Language Processing},
  pages={10925--10939},
  year={2025}
}

@article{zhu_where_2025,
	title = {Where {LLM} agents fail and how they can learn from failures},
	author = {Zhu, Kunlun and Liu, Zijia and Li, Bingxuan and Tian, Muxin and Yang, Yingxuan and Zhang, Jiaxun and Han, Pengrui and Xie, Qipeng and Cui, Fuyang and Zhang, Weijia and Ma, Xiaoteng and Yu, Xiaodong and Ramesh, Gowtham and Wu, Jialian and Liu, Zicheng and Lu, Pan and Zou, James and You, Jiaxuan},
	journal = {arXiv preprint arXiv:2509.25370},
	year = {2025},
}

@inproceedings{zhang_soft_2025,
	title={Soft thinking: Unlocking the reasoning potential of llms in continuous concept space},
  author={Zhang, Zhen and He, Xuehai and Yan, Weixiang and Shen, Ao and Zhao, Chenyang and Wang, Xin},
  booktitle={Advances in Neural Information Processing Systems},
  volume={38},
  pages={168990--169012},
  year={2025}
}

@inproceedings{hendrycks_measuring_2021,
	title={Measuring Mathematical Problem Solving With the {MATH} Dataset},
author={Dan Hendrycks and Collin Burns and Saurav Kadavath and Akul Arora and Steven Basart and Eric Tang and Dawn Song and Jacob Steinhardt},
booktitle={Thirty-fifth Conference on Neural Information Processing Systems Datasets and Benchmarks Track (Round 2)},
year={2021}
}

@inproceedings{jain_livecodebench_2025,
	title = {{LiveCodeBench}: Holistic and contamination free evaluation of large language models for code},
	author = {Jain, Naman and Han, King and Gu, Alex and Li, Wen-Ding and Yan, Fanjia and Zhang, Tianjun and Wang, Sida and Solar-Lezama, Armando and Sen, Koushik and Stoica, Ion},
	booktitle = {International Conference on Learning Representations},
	year = {2025},
}

@article{chen_towards_2025,
	title = {Towards reasoning era: A survey of long chain-of-thought for reasoning large language models},
	author = {Chen, Qiguang and Qin, Libo and Liu, Jinhao and Peng, Dengyun and Guan, Jiannan and Wang, Peng and Hu, Mengkang and Zhou, Yuhang and Gao, Te and Che, Wanxiang},
	journal = {arXiv preprint arXiv:2503.09567},
	year = {2025},
}

@article{li_system_2025,
	title = {From {System 1} to {System 2}: A survey of reasoning large language models},
	author = {Li, Zhong-Zhi and Zhang, Duzhen and Zhang, Ming-Liang and Zhang, Jiaxin and Liu, Zengyan and Yao, Yuxuan and Xu, Haotian and Zheng, Junhao and Wang, Pei-Jie and Chen, Xiuyi and Zhang, Yingying and Yin, Fei and Dong, Jiahua and Li, Zhiwei and Bi, Bao-Long and Mei, Ling-Rui and Fang, Junfeng and Liang, Xiao and Guo, Zhijiang and Song, Le and Liu, Cheng-Lin},
	journal = {arXiv preprint arXiv:2502.17419},
	year = {2025},
}

@article{sui_stop_2025,
	title = {Stop overthinking: A survey on efficient reasoning for large language models},
	author = {Sui, Yang and Chuang, Yu-Neng and Wang, Guanchu and Zhang, Jiamu and Zhang, Tianyi and Yuan, Jiayi and Liu, Hongyi and Wen, Andrew and Zhong, Shaochen and Zou, Na and Chen, Hanjie and Hu, Xia},
	journal = {Transactions on Machine Learning Research},
	year = {2025},
}


\appendix

\section{Additional Experimental Settings}\label{sec:additional-exp-settings}

\subsection{Datasets and Splits}\label{sec:dataset-construction}

\begin{table}[!htbp]
\centering
\small
\caption{Dataset partitions. MATH-500 and the LiveCodeBench test window are not used for training or model selection.}
\label{tab:dataset_splits}
\begin{tabular}{llll}
\toprule
\textbf{Dataset} & \textbf{Train} & \textbf{Validation} & \textbf{Evaluation} \\
\midrule
MATH & MATH train & held-out MATH train & MATH-500 \\
LiveCodeBench & May 2023--June 2024 & July--September 2024 & October 2024--April 2025 \\
\bottomrule
\end{tabular}
\end{table}

Training continuations, candidate steps, and all learned parameters use only the training partition in Table~\ref{tab:dataset_splits}. Validation data select checkpoints and hyperparameters; evaluation data are used only for reporting results.

\paragraph{MATH and MATH-500.} The MATH training split contains $7{,}500$ competition problems across seven subjects~\citep{hendrycks_measuring_2021}. We reserve a fixed subset for validation and use the remaining problems for both stages of \Method{} training. Evaluation uses all 500 problems in MATH-500~\citep{lightman_lets_2023}.

\paragraph{LiveCodeBench.} We use the code-generation portion of LiveCodeBench release~v6, which contains problems dated from May~2023 through April~2025~\citep{jain_livecodebench_2025}. Release dates define the chronological train, validation, and test windows in Table~\ref{tab:dataset_splits}. Evaluation uses all 341 code-generation problems released during the October~2024--April~2025 test window.

\paragraph{Reasoning format and verification.} For MATH, a reasoning step is one numbered mathematical step, and the final boxed answer is checked by normalized symbolic equivalence. For LiveCodeBench, the model produces numbered algorithm-design steps followed by executable code; a reasoning step is one numbered planning step, and the final code is checked with the benchmark test harness.

\subsection{Hyperparameters}\label{sec:hyperparameters}

Table~\ref{tab:hparams} summarizes the hyperparameters used across the two training stages and at inference. The same settings are used for MATH and LiveCodeBench unless stated otherwise, with early stopping on the corresponding validation partition.

\begin{table}[!htbp]
\centering
\footnotesize
\renewcommand{\arraystretch}{1.05}
\caption{Hyperparameters used for \Method{} training and inference.}
\label{tab:hparams}
\begin{tabular}{p{0.34\linewidth}p{0.59\linewidth}}
\toprule
\multicolumn{2}{l}{\textit{Architecture}} \\
\midrule
Backbones                         & Qwen2.5-7B-Instruct; Llama-3.1-8B-Instruct \\
LoRA rank                         & $16$ \\
LoRA $\alpha$                     & $32$ \\
LoRA dropout                      & $0.05$ \\
LoRA target modules               & $\{q,k,v,o\}\_$proj of every attention layer \\
Projection head $h_\phi$          & 2-layer MLP, hidden $1024$, GELU activation \\
Hyperbolic embedding dim $n$      & $64$ \\
Directional guidance head $q_\omega$ & predicts a tangent vector from the state and problem-root embeddings \\
Up-projector $g_\psi$             & 2-layer MLP, hidden $2048$, GELU activation \\
Up-projector output dim           & backbone hidden size ($3584$ for Qwen2.5; $4096$ for Llama-3.1) \\
Curvature $c$ init                & $1.0$ (learnable scalar) \\
Training hardware                 & $4\times$ NVIDIA H200 GPUs \\
Random seeds                      & $4$, $44$, $444$ \\
\midrule
\multicolumn{2}{l}{\textit{Stage 1: state-geometry training}} \\
\midrule
Optimizer                         & AdamW, $\beta_1{=}0.9,\ \beta_2{=}0.999,\ \epsilon{=}10^{-8}$ \\
Learning rate                     & $1\mathrm{e}{-3}$ \\
Weight decay                      & $0.01$ \\
LR schedule                       & cosine with linear warmup \\
Warmup steps                      & $500$ \\
Epochs                            & $20$ \\
Effective batch size              & $64$ ($4$ GPUs $\times$ $16$ per-GPU) \\
Ranking-loss margin               & $0.1$ \\
Gradient clipping                 & $1.0$ \\
Mixed precision                   & bf16 \\
Continuations per state $K$       & $32$ \\
Sampling temperature $\tau_{\mathrm{mc}}$ & $0.8$ \\
Nucleus $p$                       & $0.95$ \\
Max tokens per continuation       & $1024$ \\
Failure cost $C_{\mathrm{fail}}$  & one plus the longest verified training continuation (logical steps) \\
Importance-weight decay $\eta$    & $0.95$ \\
\midrule
\multicolumn{2}{l}{\textit{Stage 2: directional-guidance training}} \\
\midrule
Optimizer                         & AdamW \\
Learning rate                     & $5\mathrm{e}{-5}$ (Stage~2 trainable parameters) \\
Weight decay                      & $0.0$ \\
LR schedule                       & cosine, warmup ratio $0.03$ \\
Epochs                            & $5$ \\
Reasoning trajectories per training problem & $8$ \\
Sampling temperature $\tau$       & $0.7$ \\
Nucleus $p$                       & $0.9$ \\
Max tokens per reasoning trajectory & $1024$ \\
Guidance-preference weight $\lambda_{\Delta}$ & $1.0$ \\
Directional margin $\gamma_{\mathrm{dir}}$ & $0.1$ \\
Auxiliary-text weight $\lambda_{\mathrm{text}}$ & $0.5$ \\
Geometry-retention weight $\lambda_{\mathrm{keep}}$ & $0.1$ \\
Effective batch size              & $256$ \\
Mixed precision                   & bf16 \\
\midrule
\multicolumn{2}{l}{\textit{Inference (\Method{} and single-pass baselines)}} \\
\midrule
Decoding                          & greedy \\
Max new tokens                    & $1024$ \\
Re-encode interval                & every step boundary \\
\bottomrule
\end{tabular}
\end{table}

\section{Training and Inference Details}\label{sec:training-details}

\subsection{\Method{} Implementation}\label{sec:hyperguide-impl}

All architecture and optimization hyperparameters are listed in Table~\ref{tab:hparams}. During Stage~2 and inference, the state encoder processes the plain-text reasoning state with LoRA disabled, while the decoder uses the Stage~2 adapter. The state encoder remains trainable during Stage~2.

\subsection{Hyperbolic State Geometry}\label{app:geometry-details}

\paragraph{Distance and projection.} For the Poincar\'e ball defined in Section~\ref{sec:preliminaries}, the geodesic distance is
\begin{equation}
    d_{\mathbb D}(\mathbf x,\mathbf y)
    = \frac{1}{\sqrt c}\cosh^{-1}\!\left(
    1+\frac{2c\|\mathbf x-\mathbf y\|^2}
    {(1-c\|\mathbf x\|^2)(1-c\|\mathbf y\|^2)}
    \right).
    \label{eq:poincare_dist}
\end{equation}
Distance to the origin is $(2/\sqrt c)\tanh^{-1}(\sqrt c\|\mathbf x\|)$. The projection head maps Euclidean hidden representations into the ball using the exponential map at the origin~\citep{octavian-eugen_ganea_hyperbolic_2018}.

\paragraph{Objectives.} The radial ranking and tree-metric objectives are given in Equations~\ref{eq:stage1-rank-main} and~\ref{eq:stage1-metric-main}. Appendix~\ref{app:mc-details} describes the construction and weighting of these triplets.

\subsection{Directional Guidance Training}\label{app:direction-details}\label{app:action-aware-details}

\paragraph{Transition directions.} The state and successor embeddings define the transition directions $\mathbf u_t^{(i)}$ of Section~\ref{ssec:directional-guidance}, while the guidance head predicts a tangent vector at the current state. Gradients through these directions update the state encoder during Stage~2. The transport in Equation~\ref{eq:virtual-token} is computed in closed form on the Poincar\'e ball using gyrovector operations~\citep{octavian-eugen_ganea_hyperbolic_2018}.

\paragraph{Guidance preference.} The two likelihood margins in Equation~\ref{eq:margins} are evaluated with the same adapted decoder. Consequently, $\mathcal L_{\Delta}$ in Equation~\ref{eq:guide-loss} isolates the change in candidate preference caused by inserting the guidance token.

\paragraph{Reasoning states used in Stage~2.} Each epoch generates reasoning trajectories with the current model and constructs candidate transitions at the resulting reasoning states. Refreshing these states after each epoch keeps the guidance data aligned with the trajectories produced by the adapted decoder.

\paragraph{Continuous decoding.} Inference uses one live decoder cache for the entire reasoning sequence. The prompt is prefilled, and its root embedding is computed once. At each completed logical step, the accumulated plain-text state is encoded separately, and the resulting guidance direction is converted to one virtual token and appended as one additional cached position. Autoregressive decoding then continues without re-prefilling the prefix. Logical step completion is task-aware and does not split multiline mathematical blocks. Step count, per-step token count, and total generated tokens have separate stopping limits. No candidate expansion, scoring, or reranking is performed at inference.

\subsection{Auxiliary Text Loss}\label{app:text-loss}

Let $\mathcal D_{\mathrm{text}}$ contain pairs $(s_t,y)$ where $y$ is either a minimum-cost candidate step $a_t^{(i)}$ with $i\in\mathcal P(s_t)$ (Section~\ref{ssec:directional-guidance}) or a verified final-answer step. Candidate-step targets are taken only from states with at least one candidate that has a sampled verified continuation. The weight $\mu(y)$ allows these target types to receive different weights. With the virtual guidance token $\mathbf v_t$ of Equation~\ref{eq:virtual-token}, the token-level cross-entropy is
\begin{equation}
    \mathcal L_{\mathrm{NLL}}
    = -\mathbb E_{(s_t,y)\sim\mathcal D_{\mathrm{text}}}
    \left[
    \mu(y)\sum_{j=1}^{|y|}
    \log p\!\left(y_j\mid s_t,\mathbf v_t,y_{<j}\right)
    \right].
    \label{eq:text-nll}
\end{equation}
We retain the Stage~1 geometry while updating the state encoder with
\begin{equation}
    \mathcal L_{\mathrm{Text}}
    = \mathcal L_{\mathrm{NLL}}
    + \lambda_{\mathrm{keep}}\mathcal L_{\mathrm{Geom}}.
    \label{eq:text-loss}
\end{equation}
For a final-answer target, $y$ contains the complete required answer form.

\subsection{Continuation and Tree Construction Details}\label{app:mc-details}

This section specifies the continuation estimator used to order Stage~2 candidate transitions.

\paragraph{Sampling and success-weighted cost.} For each state $s$, let $\rho_1,\ldots,\rho_K$ be independent continuations sampled from the frozen backbone using the settings in Table~\ref{tab:hparams}. Let $b_i\in\{0,1\}$ indicate whether $\rho_i$ reaches a final answer accepted by the verifier $\mathcal V$, and let $L_i=|\rho_i|$ count logical reasoning steps. We define
\begin{equation}
    \hat d(s)
    = \frac{1}{K}\sum_{i=1}^{K}
    \left[b_iL_i+(1-b_i)C_{\mathrm{fail}}\right],
    \label{eq:mc_value}
\end{equation}
where $C_{\mathrm{fail}}$ is set to one plus the longest verified continuation in the corresponding training set. Equivalently, $\hat d(s)$ averages successful-continuation length and the failure penalty according to their empirical frequencies. Assigning the fixed penalty rather than the observed length of an unsuccessful rollout prevents an early incorrect termination from receiving a low cost. Thus, shorter successful continuations receive lower cost, while an isolated short success is penalized by the failures in the remaining samples.

\paragraph{Tree distances.} Pairwise tree distances for $\mathcal{L}_{\mathrm{metric}}$ in Equation~\ref{eq:stage1-metric-main} are extracted from generated continuations. Two types of pairs are used:
\begin{enumerate}[leftmargin=*,itemsep=2pt]
  \item \textbf{Pairs within one continuation.} Consecutive states along a single
    continuation $\rho$ satisfy $d_{\mathcal{T}}(s_t, s_{t+k}) = k$.
  \item \textbf{Pairs across continuations.} Two continuations $\rho^{(1)},\rho^{(2)}$
    that share a common prefix up to step $t$ and then diverge yield
    $d_{\mathcal{T}}\!\bigl(s^{(1)}_{t+j},\, s^{(2)}_{t+k}\bigr) = j + k$.
\end{enumerate}

\paragraph{Importance weighting.} Trajectory-local distances become less reliable as the offset from the shared prefix grows, because the inferred tree structure is based on a finite sample of continuations. To down-weight distant pairs, each triplet $(s_i, s_j, s_k)$ in $\mathcal{L}_{\mathrm{metric}}$ receives the multiplicative weight
\begin{equation}
  w_{ijk} \;=\;
  \eta^{\,d_{\mathcal{T}}(s_i,\,s_j) \,+\, d_{\mathcal{T}}(s_i,\,s_k)},
  \label{eq:iw-decay}
\end{equation}
where the decay factor $\eta$ is specified in Table~\ref{tab:hparams}. This ensures that the loss is dominated by nearby, high-confidence distance estimates while still receiving gradient from longer-range pairs.

\subsection{Baseline Implementations}\label{sec:baseline-impl}

\paragraph{Few-shot greedy.} Few-shot greedy performs one greedy decode from the task prompt.

\paragraph{Self-Consistency (MATH).} Following \citet{wang_self-consistency_2023}, we generate five reasoning trajectories at temperature $0.7$ and nucleus $p{=}0.95$ and use majority voting over normalized final answers.

\paragraph{Tree of Thoughts.} Following \citet{yao_tree_2023}, each active reasoning state is expanded with three candidate next steps and scored by the backbone using a task-specific value prompt. Beam search retains the five highest-scoring states at each depth.

\paragraph{PT-SFT.} Following \citet{wang_guiding_2024}, each reasoning step is assigned a discrete latent planning variable inferred from its contextualized representation, and the corresponding learned planning token is inserted at the beginning of the step. The model is fine-tuned to autoregressively generate each planning token before its reasoning step and the final answer; at inference, it likewise predicts a planning token before generating each step.

\paragraph{Outcome Value Model.} Following \citet{yu_ovm_2024}, a scalar head is trained on generated reasoning trajectories labeled by final-answer correctness. At inference, the head scores candidate next steps in a width-five beam search.

\paragraph{Coconut and SoftCoT.} We use the public training procedures of Coconut~\citep{hao_training_2025} and SoftCoT~\citep{xu_softcot_2025}, replacing their task data with the MATH or LiveCodeBench training partition. Their adapters are selected on the same validation data as \Method{} and decoded with the same answer format and length limit.

\paragraph{Soft Thinking.} Following \citet{zhang_soft_2025}, Soft Thinking forms continuous concept tokens from probability-weighted mixtures of token embeddings and requires no task-specific training.

\paragraph{Random seeds.} For the training-based baselines, the seed controls training and any sampling at inference; for Self-Consistency and Tree of Thoughts, it controls the sampling of trajectories and candidate steps. Few-shot and Soft Thinking decode greedily without training, so their outputs do not depend on the seed, and we report a single run.

\section{Additional Experiments}
\subsection{Additional Ablation Studies}\label{app:additional-ablation}

\begin{table}[t]
\centering
\footnotesize
\caption{Additional ablations on Qwen2.5-7B-Instruct. All conditions are trained and evaluated with seed 4.}
\label{tab:additional-ablation}
\begin{adjustbox}{max width=\linewidth}
\begin{tabular}{lcccc}
\toprule
& \multicolumn{2}{c}{MATH-500} & \multicolumn{2}{c}{LiveCodeBench} \\
\cmidrule(lr){2-3}\cmidrule(lr){4-5}
Variant & Accuracy $\uparrow$ & Tokens $\downarrow$ & Pass@1 $\uparrow$ & Tokens $\downarrow$ \\
\midrule
\textbf{\Method{}} &77.4 &352.7 &25.2 &373.5 \\
\quad Binary Stage~2 &73.2 &522.3 &20.5 &377.1 \\
\quad Ordinal-preference LoRA (w/o geometry or token) &74.8 &397.4 &22.0 &379.1 \\
\bottomrule
\end{tabular}
\end{adjustbox}
\end{table}

\textit{Binary Stage~2.} Here we replace ordinal continuation costs with binary labels that distinguish successors with and without a verified continuation. This variant has a 4.2-point accuracy drop on MATH-500 and a 4.7-point Pass@1 drop on LiveCodeBench. It also substantially increases length on MATH-500, from 352.7 to 522.3 tokens. Binary labels can steer the model away from steps with no verified continuation, but they give no preference among successors with different empirical success rates or successful-continuation lengths.

\textit{Ordinal-preference LoRA (w/o geometry or token).} This variant removes the state geometry and guidance token and instead trains the same decoder LoRA directly on the same rollout-ranked candidate pairs, increasing the likelihood margin of the lower-cost step over the higher-cost one. It reaches 74.8\% accuracy on MATH-500 and 22.0\% Pass@1 on LiveCodeBench, outperforming Binary Stage~2 by 1.6 and 1.5 points, respectively. However, it remains 2.6 points below \Method{} on MATH-500 and 3.2 points below it on LiveCodeBench, while also generating longer outputs. 



\end{document}